\documentclass{spie}

\usepackage{times} 
\usepackage{indentfirst} 

\usepackage{xkeyval}
\usepackage{graphicx}
\usepackage{subcaption}
\usepackage{hyperref}
\usepackage{multirow}

\usepackage{amsmath}
\usepackage{amssymb}

\DeclareMathOperator*{\argmin}{arg\,min}
\DeclareMathOperator{\sgn}{sgn}
\usepackage{mathtools}

\usepackage{algorithm}
\usepackage{algpseudocode}

\title{Dynamic Programming Approach to Template-based OCR}

\author{Mikhail A. Povolotskiy\supit{1, 2}, Daniil V. Tropin\supit{2, 3}
  \skiplinehalf
  \normalsize
  \supit{1} Institute for Information Transmission Problems RAS, Moscow, Russia; \\
  \supit{2} Moscow Institute of Physics and Technology “MIPT”, Dolgoprudny, Russia; \\
  \supit{3} Smart Engines, Moscow, Russia;
}

\begin{document}

\maketitle

\begin{abstract}
  In this paper we propose a dynamic programming solution to the template-based recognition task in OCR case.
  We formulate a problem of optimal position search for complex objects consisting of parts forming a sequence.
  We limit the distance between every two adjacent elements with predefined upper and lower thresholds.
  We choose the sum of penalties for each part in given position as a function to be minimized.
  We show that such a choice of restrictions allows a faster algorithm to be used than the one for the general form of deformation penalties.
  We named this algorithm Dynamic Squeezeboxes Packing (DSP) and applied it to solve the two OCR problems: text fields extraction from an image of document Visual Inspection Zone (VIZ) and license plate segmentation.
  The quality and the performance of resulting solutions were experimentally proved to meet the requirements of the state-of-the-art industrial recognition systems.

\end{abstract}

\section{Introduction}

Template-based recognition is an approach which involves recognition of composite objects on an image with the use of separate parts models and joint geometric relations models described by a template.
As a result, correspondences between model parts and image regions are built.

Template-based recognition is used in various detection and recognition tasks.
For example, in the paper \cite{felzenszwalb2010object} people, animals, vehicles and furniture are detected.
In \cite{chrysos2015offline}, face tracking-by-detection is performed.
Template-based approach can also be applied to detect contours \cite{zhang2017graph}.

Objects which can be described by templates occur in optical character recognition (OCR) problems as well.
For example, in the paper \cite{sheshkus2015approach} the information zone of a credit card is represented as a three-line structure which helps splitting it into separate lines.
Structured documents and vehicle license plates are the objects which are examined in this paper.
Parts position estimation alone is typically insufficient in OCR case as text recognition should be performed for some regions of interest.
Building correspondence between image regions and model parts prevents a lot of recognition errors if the allowed content of a certain part is a priori limited.
For example, if a document field can only contain 8 digits date, its corresponding image region must contain exactly 8 digits and no letters, and not any digits combinations are valid.
The problem of template-based recognition can be stated in a certain manner so dynamic programming can be used \cite{felzenszwalb2011dynamic}.
In particular, it is possible when parts of a structure form a sequence:
\begin{equation} \label{eq:general}
    Z(l) \equiv \sum_{i=1}^N m_i(l_i) + \sum_{i=1}^{N-1} d_{i}(l_i, l_{i+1}) \to \min_{l_1, \ldots, l_N},
\end{equation}
where $Z(l)$ is a function being minimized; $l = l_1, \ldots, l_N$ are parts locations; $m_i(l_i)$ is a cost for assigning location $l_i$ to $i^{th}$ part; $d_{i}(l_i, l_{i+1})$ is a cost for deformation of a link between adjacent parts.

The general algorithm proposed in \cite{felzenszwalb2011dynamic} has high computational complexity of $O(NW^2)$, where $N$ is a number of parts, $W$ is a number of feasible locations for one part.
We propose a more efficient algorithm named Dynamic Squeezeboxes Packing (DSP).
To allow the use of this algorithm, we require specific form of cost functions $d_{ij}(l_i, l_j)$.
We investigate DSP use for two OCR problems: text fields extraction and license plate segmentation.

\section{Dynamic Squeezeboxes Packing Algorithm}

In this section, we describe a solution to the optimization problem \eqref{eq:general} with costs for link deformations in the following form:
\begin{equation} \label{eq:deformations}
    d_{i}(l_i, l_{i+1}) =
    \begin{dcases}
        0, & T_{min}(i) \leqslant l_{i+1} - l_i \leqslant T_{max}(i), \\
        +\infty, & \text{otherwise},
    \end{dcases}
\end{equation}
where $T_{min}(i)$ and $T_{max}(i)$ are distance thresholds between $i^{th}$ and $(i+1)^{th}$ parts.
In this case, the function \eqref{eq:general} can be simplified if constrains are added:
\begin{equation} \label{eq:special}
\begin{gathered}
    Z(l) = \sum_{i=1}^N m_i(l_i) \to \min_l, \\
    T_{min}(i) \leqslant l_{i+1} - l_i \leqslant T_{max}(i), \quad i = 1, \ldots, N-1.
\end{gathered}
\end{equation}
Such distance variations remind squeezebox playing, hence the name.

The deformation model \eqref{eq:deformations} makes an investigated object impossible to recognize correctly in case of deformations exceeding the boundaries.
All the deformations within the boundaries are equally feasible and not penalized.

In the process of finding the solution two tables are being completed dynamically: the table $L(i, j)$ of $(i-1)^{th}$ part optimal locations and the table $C(i, j)$ of cumulative costs for first $i$ parts provided that $i^{th}$ part is in position $j$.
The first row of $L$ is being skipped, the one of $C$ is being copied from the values of $m_1(j)$:
\begin{equation}
    C(1, j) = m_1(j), \qquad j = 1, \ldots, W.
\end{equation}
The further procedure of tables completion is the following:
\begin{equation}
    \begin{gathered}
        L(i+1, j) = \argmin_{j - T_{max}(i) \leqslant j' \leqslant j - T_{min}(i)} C(i, j'), \\
        C(i+1, j) = C(i, L(i+1, j)) + m_{i+1}(j),
    \end{gathered}
    \qquad
    \begin{gathered}
        i = 1, \ldots N-1, \\
        j = 1, \ldots, W.
    \end{gathered}
\end{equation}
Once the tables are completed, the optimal location $l_N$ of the last part is determined as the index of the global minimum in the last row of $C$ \eqref{eq:l_N}.
The optimal locations of the remaining parts $l_1, \ldots, l_{N-1}$ are calculated through the back traverse of the $L$ \eqref{eq:l_i}.
The minimum of $Z(l)$ corresponds to the value of global minimum in the last row of $C$ \eqref{eq:Z_min}.

\begin{gather}
    \label{eq:l_N} l_N = \argmin_{1 \leqslant j \leqslant W} C(N, j), \\
    \label{eq:l_i} l_i = L(i+1, l_{i+1}), \quad i = N-1, \ldots, 1, \\
    \label{eq:Z_min} Z_{min} \equiv \min_l Z(l) = C(N, l_N).
\end{gather}

Let us estimate the algorithm computational complexity.
Single row completion for the both tables $L$ and $C$ requires $W$-fold minimum computation on a range with the fixed length of $T_{max}(i) - T_{min}(i)+ 1$ which can be calculated in $O(W)$ operations with the use of van Herk/Gil-Werman algorithm \cite{van1992fast}.
Therefore, whole tables completion needs $O(NW)$ operations.
Global minimum search needs $O(W)$, back traverse needs $O(N)$, hence, aggregate computational complexity of DSP algorithm constitutes $O(NW)$ operations which is considerably less than that of the general-case algorithm.
Notably, DSP algorithm is asymptotically optimal since its complexity is proportional to the size of the input data.

\section{Text Fields Extraction}
\begin{figure}[ht!]
  \centering
  \includegraphics[width=0.4\textwidth]{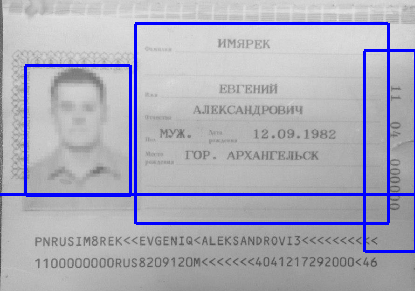}
  \caption{ Visual inspection zones are marked by rectangles}
  \label{fig:viz}
\end{figure}

For most identification documents there are large area text zones which contain basic information: full name, birth date, gender etc., hereinafter referred to as visual inspection zones (VIZ) (see Figure \ref{fig:viz}). Structure of VIZ is similar for one document type: fixed number of rows, similar distance between adjacent rows, number of text blocks in each row and distance between them. With these constrains the problem of VIZ segmentation can be decomposed, so that several stages of solution can be reduced to DSP algorithm.

\begin{figure}[ht!]
  \centering
  \begin{subfigure}[b]{0.4\linewidth}
    \includegraphics[width=\linewidth]{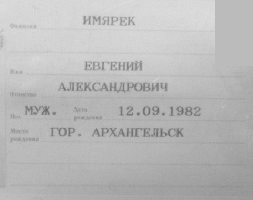}
    \caption{Input}
  \end{subfigure}
  \begin{subfigure}[b]{0.4\linewidth}
    \includegraphics[width=\linewidth]{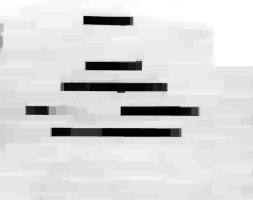}
    \caption{Output}
  \end{subfigure}
  \caption{VIZ image preprocessing}
  \label{fig:input_img}
\end{figure}

Input of the VIZ segmentation algorithm consists of greyscale normalized image of VIZ $I(x, y)$ (see Figure \ref{fig:input_img}.a) and a template with a priori information about zone structure. The template is represented by top-down sequence $\{\phi_i\}$ of rows one of two types: ``text string'' and ``gap'' alternating with each other.
For each row $\phi_i$ the minimum $h_i^{min}$ and maximum $h_i^{max}$ heights are specified. Besides that each row of type ``text string'' has its own structure: sequence of alternating blocks $\{\psi_{i, j}\}$ of two types: ``field'' and ``gap''.
For every block $\psi_{i, j}$ width limits $w_{i, j}^{min}$ and $w_{i, j}^{max}$ are known.

The output of the algorithm is expected to be a set $L = \left\{t_i, b_i, \{l_{i, j}, r_{i, j}\}_{j = 1}^{M_i}\right\}_{i = 1}^N$ of upper and lower borders $[t_i, b_i)$ for every horizontal row $\phi_i$ as well as left and right borders $[l_{i, j}, r_{i, j})$ for each block $\psi_{i, j}$, so that blocks of type ``field'' cover physical text with minimum area of background.
It is required that the rows must not intersect and must cover the image completely.
Hence the following constraints:

\begin{itemize}
    \item the height of the rows and the width of the blocks lie within the ranges specified in the template
    \begin{equation} \label{eq:restriction_1}
        \begin{gathered}
            h_i^{min} \leqslant b_i - t_i \leqslant h_i^{max}, \qquad i = 1, \ldots, N, \\
            w_{i, j}^{min} \leqslant r_{i, j} - l_{i, j} \leqslant w_{i, j}^{max}, \qquad j = 1, \ldots, M_i, \qquad i = 1, \ldots, N,
        \end{gathered}
    \end{equation}
    where $N$ is the number of rows, $M_i$ is the number of blocks inside the $i^{th}$ row;
    \item there are no gaps between adjacent rows and blocks
    \begin{equation} \label{eq:restriction_2}
        \begin{gathered}
            b_i = t_{i+1}, \qquad i = 1, \ldots, N - 1, \\
            r_{i, j} = l_{i, j + 1}, \qquad j = 1, \ldots, M_i - 1, \qquad i = 1, \ldots, N;
        \end{gathered}
    \end{equation}
    \item the boundaries of the extreme rows and blocks are equal to the boundaries of the image:
    \begin{equation} \label{eq:restriction_3}
        \begin{gathered}
            t_1 = 0, \quad b_N = H, \\
            l_{i, 1} = 0, \quad r_{i, M_i} = W, \qquad i = 1, \ldots, N,
        \end{gathered}
    \end{equation}
    where $W$ and $H$ are the width and the height of the image respectively.
\end{itemize}

The constraints \eqref{eq:restriction_2} and \eqref{eq:restriction_3} allow to assign the position of each row or block as a single number instead of two.
For example $t_i \equiv y_i$ for rows and $l_{i, j} \equiv x_{i, j}$ for blocks.
The remaining boundaries are determined trivially.

Let us assume that the average brightness of the text in the document image is less than the average background brightness.
The more text and the less background are inside the fields the better the segmentation result is.
Therefore, for the optimization problem it is natural to choose a function, which is optimal at high average brightness of pixels outside the fields and low one inside.
An example of such function is inter-class dispersion \cite{otsu1979threshold}, multiplied by the sign of the difference in the average brightness of the classes:
\begin{equation} \label{eq:otsu}
    V = \omega_0 \ \omega_1 (\mu_0 - \mu_1)^2 \sgn(\mu_0 - \mu_1) = \omega_0 \ \omega_1 (\mu_0 - \mu_1) |\mu_0 - \mu_1| \to \max,
\end{equation}
where $\omega_0$ and $\omega_1$ are fractions of the zero and the first classes from all the image pixels, $\mu_0$ and $\mu_1$ are mean intra-class brightness values.
The pixels inside the ``gaps'' belong to the zero class.
The pixels inside the ``fields'' belong to the first class.

To increase the inter-class contrast, we apply a number of image processing operations (see Figure \ref{fig:input_img}).
Those operations are: morphological closing with the quadratic structuring element $11\times11$, subtraction of original image from closing result, morphological opening of inverted subtraction result with horizontal structuring element $31\times1$, morphological closing of the previous result with the vertical structuring element $1\times7$, and auto contrast operation.
The constants above are chosen for text height of 9 pixels and serve as an example.

The function \eqref{eq:otsu} can not be represented as a sum of single part costs.
However, if we assume that the row heights $h_{2k}$ and the field widths $w_{2k, 2m}$ are fixed, then the maximization of V is equivalent to minimizing the total brightness of the pixels within the ``fields''.

To prove this, let us rewrite \eqref{eq:otsu} in terms of total brightness $S_0$ and $S_1$ and volumes $Q_0$ and $Q_1$ of classes:
\begin{equation} \label{eq:otsu_rewritten} \begin{gathered}
    \omega_0 = \frac{Q_0}{Q_0 + Q_1}, \qquad \omega_1 = \frac{Q_1}{Q_0 + Q_1}, \qquad \mu_0 = \frac{S_0}{Q_0}, \qquad \mu_1 = \frac{S_1}{Q_1}, \\
    V = \frac{Q_0 Q_1}{(Q_0 + Q_1)^2} \left( \frac{S_0}{Q_0} - \frac{S_1}{Q_1} \right) \left| \frac{S_0}{Q_0} - \frac{S_1}{Q_1} \right| \to \max.
\end{gathered} \end{equation}
It is obvious that the value $S = S_0 + S_1$ is constant and equal to the sum of the brightness of all the pixels of the image $I(x, y)$.
Let us express $S_0$ through it in the formula \eqref{eq:otsu_rewritten}:
\begin{equation} \label{eq:otsu_final} \begin{aligned}
    V & = \frac{Q_0 Q_1}{(Q_0 + Q_1)^2} \left( \frac{S - S_1}{Q_0} - \frac{S_1}{Q_1} \right) \left| \frac{S - S_1}{Q_0} - \frac{S_1}{Q_1} \right| = \\
    & = \frac{Q_1}{Q_0 (Q_0 + Q_1)^2} \left( S - \frac{Q_0 + Q_1}{Q_1} S_1 \right) \left| S - \frac{Q_0 + Q_1}{Q_1} S_1 \right| \to \max.
\end{aligned} \end{equation}
The volumes of classes $Q_0$ and $Q_1$ do not change after fixing sizes of fields, so $V$ is a function of a single variable $S_1$.
$V$ monotonically decreases with increasing $S_1$, therefore, minimizing $S_1$ leads to maximizing $V$, QED.

Thus, we obtain the following optimization task:
\begin{equation} \label{eq:field_sum}
    S_1 = \sum_{k = 1}^{\lfloor N/2 \rfloor} \sum_{m = 1}^{\lfloor M_{2k} / 2 \rfloor} \sum_{x = x_{2k,2m}}^{x_{2k,2m} + w_{2k,2m}} \sum_{y=y_{2k}}^{y_{2k} + h_{2k}} I(x, y) = \sum_{k = 1}^{\lfloor N/2 \rfloor} \sum_{m = 1}^{\lfloor M_{2k} / 2 \rfloor} I_{\Sigma}(x_{2k,2m}, y_{2k}, w_{2k,2m}, h_{2k}) \to \min,
\end{equation}
where $I_{\Sigma}(x, y, w, h)$ is total brightness in the window with the upper left corner at the point $(x, y)$, width $w$ and height $h$.
Note that its value can be computed for any valid values of the arguments for the constant number of operations if we first calculate the integral image \cite{viola2001rapid} for $O(WH)$ operations.

Suppose that for any position $y$ of $k^{th}$ row the minimum total brightness of its fields is known:
\begin{equation} \label{eq:one_line}
    S_k(y) = \min_{x} \sum_{m = 1}^{\lfloor M_{2k} / 2 \rfloor} I_{\Sigma}(x_{2k,2m}, y, w_{2k,2m}, h_{2k}).
\end{equation}
Then the problem  \eqref{eq:field_sum} reduces to
\begin{equation}
    \sum_{k = 1}^{\lfloor N/2 \rfloor} S_k(y_{2k}) \to \min_y.
\end{equation}
Taking into account \eqref{eq:restriction_1} the values $y$ are constrained within
\begin{equation}
    \begin{gathered}
        h_{2k} + h_{2k+1}^{min} \leqslant y_{2k+2} - y_{2k} \leqslant h_{2k} + h_{2k+1}^{max}, \qquad k = 1, \ldots, \left\lfloor \frac{N - 1}{2} \right\rfloor.
    \end{gathered}
\end{equation}
Thus the task meets the form \eqref{eq:special}, that allows us to apply DSP algorithm to solve it.
Restrictions on the positions of the first and last rows can be set by infinite penalties:
\begin{equation}
\begin{aligned}
    S_{2}'(y) = &
    \begin{dcases}
        S_2(y), & h_1^{min} \leqslant y \leqslant h_1^{max}, \\
        +\infty, & \text{else};
    \end{dcases} \\
    S_{N-1}'(y) = &
    \begin{dcases}
        S_{N-1}(y), & H - h_N^{max} \leqslant y + h_{N-1} \leqslant H - h_N^{min}, \\
        +\infty, & \text{else}.
    \end{dcases} \\
\end{aligned}
\end{equation}

\begin{figure}[h!]
  \centering
  \begin{subfigure}[t]{0.3\linewidth}
    \includegraphics[width=\linewidth]{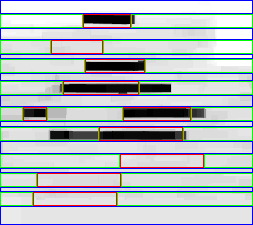}
    \caption{Result of minimizing $S_1$ with static sizes of fields using DSP}
  \end{subfigure}
  \begin{subfigure}[t]{0.3\linewidth}
    \includegraphics[width=\linewidth]{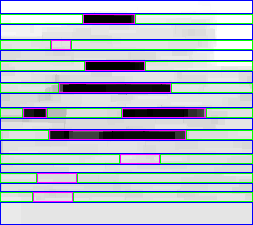}
    \caption{Result of maximizing \eqref{eq:otsu} using coordinate descent}
  \end{subfigure}
  \begin{subfigure}[t]{0.3\linewidth}
    \includegraphics[width=\linewidth]{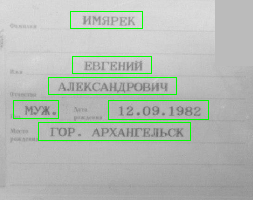}
    \caption{Result of VIZ segmentation}
  \end{subfigure}
  \caption{Stages of VIZ segmentation}
  \label{fig:segmentation_results}
\end{figure}

The problem \eqref{eq:one_line} of calculating the optimal segmentation of fields within a single row in a current position reduces to the problem \eqref{eq:special} in a similar way.
Hence, the problem \eqref{eq:field_sum} of segmenting rows and fields of fixed sizes can be solved by DSP in two stages: search for the optimal position of the fields for each possible position of each row, and then search for the optimal position of the lines themselves (see Figure \ref{fig:segmentation_results}.a).

Let us recall that according to the initial problem statement, the sizes of the fields can vary within the boundaries set by the template.
To refine the size and position of the fields, it is proposed to maximize \eqref{eq:otsu} using coordinate descent: alternately looking over the fields and their boundaries within template boundaries, increasing the target value of the function.
As an initial approximation we use the dynamic programming solution.
The stopping criterion will be considered as idle iteration search without increasing the value of the function, or reaching the maximal number of iterations given as a parameter of the algorithm (see Figure \ref{fig:segmentation_results}.b).
Final extracted fields are represented on Figure \ref{fig:segmentation_results}.c.

The total complexity of the algorithm is $O(WH\sum_{k=1}^{\lfloor N/2 \rfloor}M_{2k})$ (the complexity of the integral image is $O(WH)$, dynamic programming within rows is $O(WH\sum_{k=1}^{\lfloor N/2 \rfloor}M_{2k})$, dynamic programming on the rows is $O(HN)$, coordinate descent is $O((W+H)\sum_{k=1}^{\lfloor N/2 \rfloor}M_{2k})$).

\section{License Plate Segmentation}
In many countries, license plates follow specific format: there is one or more plate types, with each of them having fixed symbols locations.
Provided that the plate type is known, plate geometry can be described by a template so template-based approach can be applied to the plate segmentation task.

The dynamic programming solution was proposed in the previous work \cite{povolotskiy2017russian}.
Let $I(x, y)$ be the input plate image.
As a result, bounding rectangles $p_i = (x_i, y_i, w_i, h_i)$ of image regions containing symbols are expected.
As a functional to be optimized, the total brightness of pixels inside these rectangles is used:
\begin{equation}\begin{gathered}\label{eq4}
    \sum_{i=1}^{N} \sum_{x,y \in p_i} I(x, y) = \sum_{i=1}^{N} I_{\Sigma}(p_i) \to \min,
\end{gathered}\end{equation}
where $N$ is the number of symbols.
The mean brightness of symbols is supposed to be lower than that of the background; if it is incorrect for a considered plate type, the input image must be inverted.
In contrast with VIZ fields extraction, morphological filtering is not applicable in this task because gaps between symbols and background areas inside bounding rectangles are comparable in size.
However, auto contrast operation can still be applied to increase the contrast between the symbols and the background.

\begin{figure}[!ht]
  \centering
  \begin{subfigure}[t]{0.45\linewidth}
    \includegraphics[width=\linewidth]{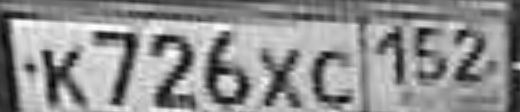}
    \caption{Source plate image}
  \end{subfigure}
  \begin{subfigure}[t]{0.45\linewidth}
    \includegraphics[width=\linewidth]{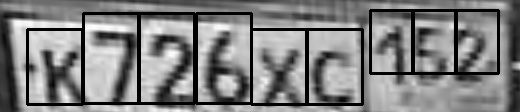}
    \caption{Iteration 0 (initial location values)}
  \end{subfigure}
  \begin{subfigure}[t]{0.45\linewidth}
    \includegraphics[width=\linewidth]{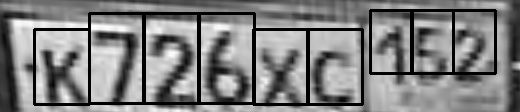}
    \caption{Iteration 1}
  \end{subfigure}
  \begin{subfigure}[t]{0.45\linewidth}
    \includegraphics[width=\linewidth]{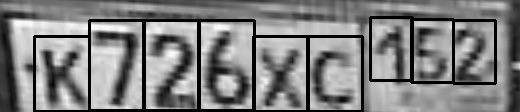}
    \caption{Iteration 2}
  \end{subfigure}
  \begin{subfigure}[t]{0.45\linewidth}
    \includegraphics[width=\linewidth]{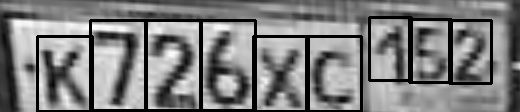}
    \caption{Iteration 3}
  \end{subfigure}
  \begin{subfigure}[t]{0.45\linewidth}
    \includegraphics[width=\linewidth]{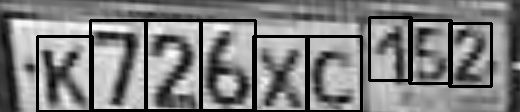}
    \caption{Iteration 4}
  \end{subfigure}
    \caption{Iterative plate symbols regions adjustment}
    \label{fig:iterations}
\end{figure}

Let us define initial symbol regions locations as $p_i^0 = (x_i^0, y_i^0, w_i^0, h_i^0)$ in the template and put the following constraints on their changes:
\begin{itemize}
    \item symbol regions do not exceed image bounds:
    \begin{equation}
        0 \leqslant x_i \leqslant W - w_i, \qquad 0 \leqslant y_i \leqslant H - h_i, \qquad i = 1, \ldots, N;
    \end{equation}
    \item maximal change in distance between adjacent symbol regions is proportional to the initial distance between their centers:
    \begin{equation}
    \begin{gathered}
        |x_{i+1} - x_i| \leqslant \delta ||c_{i+1}^0 - c_i^0||_2, \qquad |y_{i+1} - y_i| \leqslant \delta ||c_{i+1}^0 - c_i^0||_2, \\
        c_i^0 = (x_i^0 + w_i^0/2, y_i^0 + h_i^0/2), \qquad i = 1, \ldots, N;
    \end{gathered}
    \end{equation}
    \item regions sizes are constant:
    \begin{equation}
        w_i = w_i^0, \qquad h_i = h_i^0, \qquad i = 1, \ldots, N.
    \end{equation}
\end{itemize}
The resulting problem can be reduced to the form \eqref{eq:special} if one of the coordinates is fixed for all symbol regions.
This allows application of DSP algorithm to optimize $x_i$ and $y_i$ alternately.
The iterative process of symbol regions adjustment is shown on Figure \ref{fig:iterations}.

Computational complexity of the resulting algorithm is $O(WH)$ with image integral calculation being the most time-consuming subtask as optimizations of $x$ and $y$ need $O(NW)$ and $O(NH)$ operations respectively, and the number of iterations is a small constant.

\section{Experimental Results}

Both applications of DSP algorithm were implemented into industrial recognition systems: the module for simplifying entry of identity document data Smart ID Reader \cite{smartidreader} and the automatic number plate recognition module MARINA \cite{marina}.
In order to test proposed algorithms in real conditions, the experiments were carried out to estimate the contribution of segmentation subsystems into the total number of errors and time consumption.

VIZ fields extraction was performed on a set of 7700 internal Russian passports photos.
Each photo contained two pages, the one with issuing information and the one with photo, personal information, and MRZ.
Fields extraction was run on an each page separately.
We compared our approach with the template-free one, based on histogram analysis \cite{slugin2017text}.
With the use of our algorithm, the number of incorrectly recognized passports decreased by 12\% from 1094 to 958.
The number of errors caused by wrong VIZ segmentation was reduced by 38\% from 476 to 297.
The difference in numbers can be explained with the new errors produced by the following subsystems after the segmentation errors were fixed.
In relation to the total number of errors, the share of segmentation errors went down from 44\% to 31\%.

\begin{figure}[ht!]
  \centering
  \begin{subfigure}[t]{0.3\linewidth}
    \includegraphics[width=\linewidth]{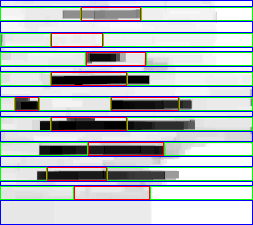}
    \caption{Layout of fields after DSP}
  \end{subfigure}
  \begin{subfigure}[t]{0.3\linewidth}
    \includegraphics[width=\linewidth]{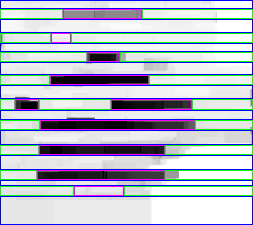}
    \caption{Layout of fields after first iteration of coordinate descent}
  \end{subfigure}
  \begin{subfigure}[t]{0.3\linewidth}
    \includegraphics[width=\linewidth]{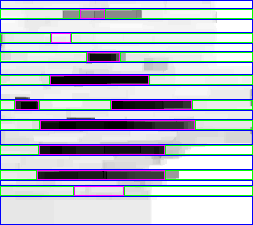}
    \caption{\centering Layout of fields after second iteration of coordinate descent}
  \end{subfigure}
  \caption{Example of coordinate descent degradation}
  \label{fig:bad_behavior}
\end{figure}

The table \ref{tab:results} shows comparison of recognition quality for different fields between histogram-based method and our approach with the different number of refinement steps.
It was observed that idle iteration tends to happen after 2-3 steps, and the optimal number of iterations is 1.
The explanation for this is following.
In practice, there are some fields-outliers with a higher mean brightness (see Figure \ref{fig:bad_behavior}).
Initially, the area of background inside coarse fields regions is big, so is the mean brightness.
Hence, pixels of fields-outliers are classified correctly, and its borders are refined well on the first iteration.
Later on, the mean brightness inside fields regions descends which leads to fields-outliers classified as background.
To fix this problem, more complicated text model is needed.

\begin{table}[!ht]
\centering
\small
\caption{Text recognition quality for different fields}
\label{tab:results}
\begin{tabular}{|c|c|c|c|c|c|c|c|c|c|}
    \hline
    Method & Authority & Authority & Birthdate & Birthplace & Gender & Issue & First & Patronymic & Surname \\
    && code &&&& date & name && \\
    \hline
    Slugin et al. & 55,34\% & 94,1\% & 90,92\% & 68,62\% & 93,57\% & 92,63\% & 91,96\% & 90,94\% & 87,34\% \\
    \hline
    DSP, 1 step & 55,37\% & 94,48\% & 93,09\% & 68,7\% & 95,58\% & 93,63\% & 94,29\% & 93,36\% & 89,49\% \\
    \hline
    DSP, $> 3$ steps & 55,24\% & 94,38\% & 92,96\% & 68,66\% & 95,54\% & 93,52\% & 94,27\% & 93,3\% & 89,2\% \\
    \hline
\end{tabular}
\end{table}

It was also measured, that fields extraction consumes 11.5\% of all Smart ID Reader computation time which is acceptable for a single subsystem of a complex program module.

The testing of license plate segmentation was performed on a data set of approximately 8000 Russian plate images.
The fraction of errors caused by segmentation subsystem comprised approximately 26\% of all MARINA errors.
A more thorough quality analysis of plate segmentation algorithm can be found in \cite{povolotskiy2017russian} and \cite{povolotskiy2018segmentation}.
Computational time took roughly 15\% of the whole MARINA run-time.

\section{Conclusion}

We proposed the Dynamic Squeezebox Packing algorithm for template-based recognition of structured objects on an image.
We applied this method to the problems of text fields extraction and license plate segmentation.
The experiments have shown that our approach allows to build fast high-quality industrial OCR systems.
As a future work, the applicability of DSP algorithm to recognize structures with non-sequential templates and different forms of deformation cost function can be investigated.
Another interesting question is a search for more complex forms of the cost function for each textual nature part which can prevent errors such as loosing text fields of higher mean brightness.

\acknowledgements

This work is supported by Russian Foundation for Basic Research (project 17-29-03236).

\bibliographystyle{spiebib}
\bibliography{bibliography}

\begin{thebibliography}{10}

\bibitem{felzenszwalb2010object}
P.~F. Felzenszwalb, R.~B. Girshick, D. McAllester, and D. Ramanan, ``Object
  detection with discriminatively trained part-based models,'' {IEEE
  transactions on pattern analysis and machine intelligence}~{\bf 32}(9),
  1627--1645 (2010).

\bibitem{chrysos2015offline}
G.~G. Chrysos, E. Antonakos, S. Zafeiriou, and P. Snape, ``Offline deformable
  face tracking in arbitrary videos,'' in Proceedings of the IEEE International
  Conference on Computer Vision Workshops{\nolinebreak\hspace{0.1em}},   1--9
  (2015).

\bibitem{zhang2017graph}
L. Zhang, H. Kong, S. Liu, T. Wang, S. Chen, and M. Sonka, ``Graph-based
  segmentation of abnormal nuclei in cervical cytology,'' {Computerized Medical
  Imaging and Graphics}~{\bf 56},  38--48 (2017).

\bibitem{sheshkus2015approach}
A. Sheshkus, D.~P. Nickolaev, A. Ingacheva, and N. Skoruykina, ``Approach to
  recognition of flexible form for credit card expiration date recognition as
  example,'' in ICMV 2015{\nolinebreak\hspace{0.1em}},   {\bf 9875}(98750R),
  1--5, SPIE (2015).

\bibitem{felzenszwalb2011dynamic}
P.~F. Felzenszwalb and R. Zabih, ``Dynamic programming and graph algorithms in
  computer vision,'' {IEEE transactions on pattern analysis and machine
  intelligence}~{\bf 33}(4),  721--740 (2011).

\bibitem{van1992fast}
M. Van~Herk, ``A fast algorithm for local minimum and maximum filters on
  rectangular and octagonal kernels,'' {Pattern Recognition Letters}~{\bf
  13}(7),  517--521 (1992).

\bibitem{otsu1979threshold}
N. Otsu, ``A threshold selection method from gray-level histograms,'' {IEEE
  transactions on systems, man, and cybernetics}~{\bf 9}(1),  62--66 (1979).

\bibitem{viola2001rapid}
P. Viola and M. Jones, ``Rapid object detection using a boosted cascade of
  simple features,'' in Computer Vision and Pattern Recognition, 2001. CVPR
  2001. Proceedings of the 2001 IEEE Computer Society Conference
  on{\nolinebreak\hspace{0.1em}},   {\bf 1},  I--I, IEEE (2001).

\bibitem{povolotskiy2017russian}
M.~A. Povolotskiy, E.~G. Kuznetsova, and T.~M. Khanipov, ``Russian license
  plate segmentation based on dynamic time warping,'' in European Conference on
  Modelling and Simulation{\nolinebreak\hspace{0.1em}},   285--291 (2017).

\bibitem{smartidreader}
 {Smart ID Reader}. \url{ http://smartengines.com/smart-id-reader/} (2018).
\newblock [Online; accessed 06-September-2018].

\bibitem{marina}
 {MARINA}. \url{ http://visillect.com/en/alpr/} (2018).
\newblock [Online; accessed 06-September-2018].

\bibitem{slugin2017text}
D.~G. Slugin and V.~V. Arlazarov, ``Text fields extraction based on image
  processing,'' {Trudy ISA RAN}~{\bf 67}(4),  65--73 (2017).

\bibitem{povolotskiy2018segmentation}
M.~A. Povolotskiy, E.~G. Kuznetsova, N.~V. Utkin, and D.~P. Nikolaev,
  ``Segmentation of vehicle registration plates based on dynamic time
  warping,'' {Sensory systems}~{\bf 32}(1),  50--59 (2018).

\end{thebibliography}

\end{document}